\documentclass{article}
\usepackage{microtype}
\usepackage{graphicx}
\usepackage{booktabs}
\usepackage{amsmath,amssymb}
\usepackage{tikz}
\usetikzlibrary{positioning,arrows.meta}
\usepackage{pgfplots}
\pgfplotsset{compat=1.16}
\usepackage{enumitem}
\setlist{nosep,leftmargin=1.4em}
\usepackage{hyperref}

\usepackage[accepted]{mlsys2025}
\mlsystitlerunning{Triage: Reliability- and Load-Conditioned Escalation for Two-Tier LLM Serving}
\newcommand{\pp}{\,pp}

\begin{document}

\twocolumn[
\mlsystitle{Triage: Reliability- and Load-Conditioned Escalation\\for Two-Tier LLM Serving with a Semantic Evidence Judge}

\begin{center}
\begin{tabular}{c@{\hspace{4em}}c}
\textbf{Basavaraj R M} & \textbf{Aryan Bharat Kumar} \\
Independent Researcher & Independent Researcher \\
\texttt{basavarajrm972@gmail.com} & \texttt{ar22073yan@gmail.com}
\end{tabular}
\end{center}

\vskip 0.3in

\begin{abstract}
Two-tier LLM serving escalates selected requests from a small model to a large one, but prompt-based routers cannot see whether retrieved evidence supports an answer, and they treat the GPU cost of an escalation as constant. We study \textsc{Triage}, which conditions escalation on a post-generation reliability state and on a load-dependent GPU price, and we report both what held and what did not on one A40 serving Qwen2.5-7B and 32B (AWQ) under vLLM. Escalation cost rose 21.9$\times$ from 1 to 10 requests/s but showed no monotone dependence on KV-cache headroom, and a pre-registered claim that the reliability state detects corrupted evidence (AUROC $\geq 0.70$) was falsified at 0.571, although any prompt-only router scores exactly 0.500 by construction. Post hoc, a semantic evidence judge raised detection to 0.846 (0.926 on never-served items) and predicted wrong small-model answers at AUROC 0.90--0.96 on RGB and three RouterBench model pairs; the resulting cascade stayed within about 3\pp\ of GPT-4 at 79.5\% lower API cost for GPT-3.5$\to$GPT-4, yet fell below random routing for open-model pairs with 2--3$\times$ price ratios. End to end, the 7B$\to$32B pair showed only a $+2.88$\pp\ held-out gap, so the allocator rightly declined to escalate and the headline is a null that a small pre-launch gate failed to expose.
\end{abstract}
]

{\let\thefootnote\relax\footnotetext{\textit{Technical report.}}}

\section{Introduction}
Serving a large language model (LLM) is expensive, and for many requests a smaller model is already correct. A common remedy is a two-tier system: a small model answers first and a large model is used only when it is likely to help \citep{chen2023frugalgpt,jitkrittum2023deferral,ding2024hybrid,hu2024routerbench,ong2025routellm}. Most deployed routers decide from the prompt alone. Two properties of real deployments make that insufficient.

First, \emph{the information that decides correctness may not be in the prompt.} In retrieval-augmented generation (RAG), the evidence attached to a question can be on-topic yet fail to contain the answer. A router that reads only the question sees an identical input whether the evidence is good or bad.

Second, \emph{the price of an escalation is not constant.} A large-model request competes with the small tier for compute. If that competition grows with load, the right escalation threshold under peak traffic differs from the one at idle.

This paper studies \textsc{Triage}, a controller that conditions on both: a reliability state (token-level uncertainty, self-consistency, evidence checks) and a load-dependent price. It chooses among stopping, probing, verifying, escalating, and abstaining. The study was run as a pre-registered GPU campaign with a deliberately adversarial reporting rule: negative and falsified results are kept in the paper.

\textbf{Contributions.}
\begin{enumerate}
\item \emph{An identification result.} For a paired evidence-corruption benchmark, every function of the question alone has AUROC exactly 0.5 (Proposition~1), verified on 400/400 text pairs and on HallusionBench.
\item \emph{A falsified pre-registered prediction and its post-hoc repair.} The inherited lexical reliability state detects RGB corruption at only 0.571. A semantic evidence judge (Jev) reaches 0.846 (0.926 held-out). We label this as a new hypothesis, not a rescue.
\item \emph{A measured load price.} Escalation cost on one A40 rises 21.9$\times$ from 1 to 10 rps; the KV-headroom dependence is a replicated null.
\item \emph{Cross-pair evidence on real outcomes.} Three RouterBench pairs, with doubly robust off-policy estimates validated against ground truth, and a multimodal test on HallusionBench.
\item \emph{An honest root-cause analysis.} The tier gap on held-out data is small, which explains the end-to-end null, and a 66-item gate failed to reveal this.
\end{enumerate}

\section{Related Work}
\textbf{Cascades and routers.} FrugalGPT cascades models by answer-quality scores \citep{chen2023frugalgpt}. \citet{ding2024hybrid} route on predicted query difficulty, and \citet{ong2025routellm} train routers from preference data. \citet{jitkrittum2023deferral} analyze when confidence-based deferral suffices. RouterBench provides a table of responses and API costs for many models on every prompt, enabling counterfactual evaluation \citep{hu2024routerbench}. Our prompt-only baselines follow this family; our difference is the post-generation reliability state and the load-dependent price.

\textbf{LLM serving.} vLLM's PagedAttention manages KV-cache memory in blocks \citep{kwon2023pagedattention}; the Splitwise study released production inference traces \citep{patel2024splitwise}. We use vLLM as the substrate and ask a systems question of it: how does the cost of one more request to a second model depend on load and cache headroom?

\textbf{Evidence quality and hallucination.} RGB measures noise robustness and negative rejection in RAG using human-collected positive and negative passages \citep{chen2024rgb}. HallusionBench pairs an image with human-edited copies and the same question \citep{guan2024hallusionbench}. Both give a paired construction in which the text of the question is identical across a ``supported''/``unsupported'' label, which is exactly what our identification result needs.

\textbf{Evaluation statistics.} We use two one-sided tests (TOST) for non-inferiority \citep{schuirmann1987tost}, cluster bootstrap \citep{efron1993bootstrap}, and inverse-propensity, self-normalized, and doubly robust estimators \citep{dudik2011dr} for off-policy evaluation (OPE).

\section{System and Method}
\subsection{Two-tier substrate and allocator}
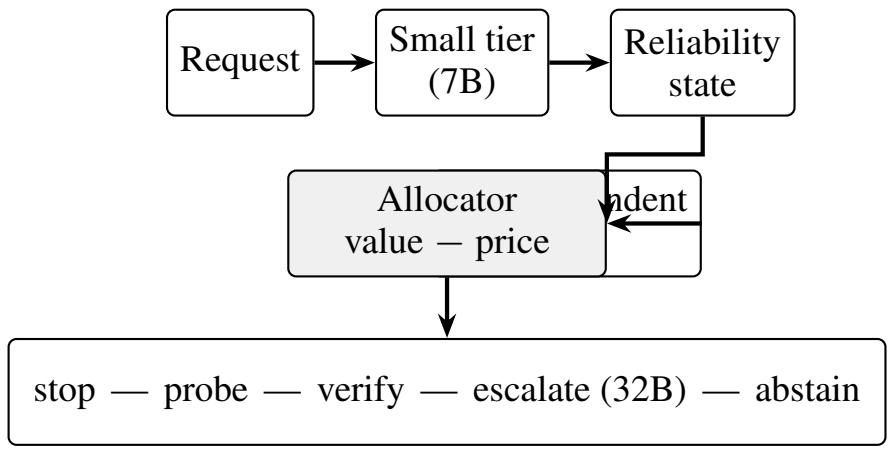
\begin{figure}[t]
\centering
\begin{tikzpicture}[
  node distance=3.5mm and 4mm,
  box/.style={draw,rounded corners=1.5pt,minimum height=7mm,align=center,font=\scriptsize,inner sep=2.5pt},
  arr/.style={-{Stealth[length=1.6mm]},thick}]
\node[box] (req) {Request};
\node[box,right=of req] (small) {Small tier\\(7B)};
\node[box,right=of small] (state) {Reliability\\state};
\node[box,below=of state,xshift=-9mm] (price) {Load-dependent\\price};
\node[box,below=of small,xshift=-1mm,fill=black!6,minimum width=2.1cm] (alloc) {Allocator\\value $-$ price};
\node[box,below=of alloc,yshift=-0.5mm,minimum width=5.8cm] (act) {stop \,|\, probe \,|\, verify \,|\, escalate (32B) \,|\, abstain};
\draw[arr] (req)--(small);
\draw[arr] (small)--(state);
\draw[arr] (state.south)--++(0,-2.5mm)-|(alloc.east);
\draw[arr] (price)--(alloc);
\draw[arr] (alloc)--(act);
\end{tikzpicture}
\caption{Control loop. The reliability state is computed after the small model answers; the price multiplier tracks measured concurrency.}
\label{fig:arch}
\end{figure}

Fig.~\ref{fig:arch} shows the control loop. The small tier answers first. A reliability state is computed from signals and passed to an allocator that compares the \emph{value} of each of eight priced interventions against their \emph{price} in GPU-seconds. The price is multiplied by a state-dependent factor that increases with measured device concurrency. The two tiers run as separate \texttt{vllm serve} processes with disjoint KV pools on one GPU.

\subsection{Reliability state}
The inherited state combines five components with fixed weights: token-level uncertainty, instability across resamples, contradiction, retrieval disagreement, and evidence sufficiency. The last is a lexical cosine score. Thresholds from an earlier API-based system were found to be unreachable for the Qwen tier: the 7B model's uncertainty on the tune split spans $[0.0033,0.1483]$ while the inherited escalation threshold was 0.28, so that branch could never fire. We recalibrate to measured percentiles.

\subsection{Identification of evidence corruption}
Let a benchmark contain, for each base item, a clean copy and a corrupted copy that share the question $q$ byte for byte and differ only in the attached evidence.

\textbf{Proposition 1.} If the positive (corrupted) and negative (clean) classes contain the same multiset of questions, any score $s(q)$ that depends on the question alone has AUROC exactly $1/2$.

\emph{Proof.} The two classes induce the same distribution on $s$. For independent draws $S^+$, $S^-$ from one distribution, $\Pr(S^+>S^-)=\Pr(S^->S^+)$, so $\Pr(S^+>S^-)+\tfrac12\Pr(S^+=S^-)=\tfrac12$. $\square$

This is a construction, not a measurement. The corruption benchmarks assert byte-identical prompts, and the check passed on 40/40 synthetic pairs (maximum absolute difference $0$), on 400/400 RGB pairs, and on HallusionBench by design. Detecting corruption therefore requires reading the evidence.

\subsection{Jev as a semantic judge in four places}
Jev (model \texttt{jev-1.13.0}, pinned) is a hosted judge that answers a structured question about a piece of text or an image \citep{typesafe2026jev}. We use it in four roles:
\begin{enumerate}
\item \emph{Evidence check} (\textit{sufficient}): given question and passages, is there enough information to answer? This replaces the cosine score and needs no answer, so it can run concurrently with generation.
\item \emph{Answer check} (\textit{relation}): given question, passages, and the served answer, does the evidence support, contradict, or say nothing about it? ``Not supported'' is used as a wrong-answer predictor.
\item \emph{Routing}: on RouterBench, a prompt-only policy has Jev read the question and routes before generation; a cascade policy has Jev read the small model's answer.
\item \emph{Image check}: given an image, question, and answer, does the image still support the answer?
\end{enumerate}
Jev is a network call, not a GPU tier; its latency (p50 356--418 ms, p95 447--806 ms) and dollar cost are reported and charged to any policy that uses it.

\section{Experimental Setup}
\textbf{Hardware and models.} One NVIDIA A40 (46,068 MiB), vLLM 0.6.3.post1, torch 2.4.0+cu121. Tiers: Qwen2.5-7B-Instruct-AWQ (small), 14B-AWQ (mid, one gate), and 32B-AWQ (big) \citep{qwen2024qwen25,lin2024awq}. At memory fraction 0.56 the big tier allocated 1,058/1,060/1,058 KV blocks on three rented hosts, and the small tier 4,355--4,358; the generation cap was 512 tokens. Client noise floor: p50 TTFT 31.1 ms, p95 45.6 ms.

\textbf{Data.} Primary mix: GPQA-diamond, MATH-500 (levels 4--5), MMLU-Pro, HotpotQA \citep{rein2023gpqa,wang2024mmlupro,yang2018hotpotqa}. Splits are by item hash (30\% tune), with a write-once record of the first test read. Corruption: RGB, 300 items, 200 served as 600 benchmark rows (corruption rates 0, 0.5, 1.0; 1,160 passage swaps) and items 200--299 held out and never served \citep{chen2024rgb}. External: RouterBench, 2,427 items from five task families, test split 1,708 \citep{hu2024routerbench}; HallusionBench, 743 images in 356 original/edited pairs \citep{guan2024hallusionbench}. GSM8K is excluded from RouterBench because its scores are quarter-point and non-exact-match, which inverted the signal.

\textbf{Statistics.} Cluster bootstrap by item (B=2,000 for AUROC; 1,000 paired for RouterBench), TOST with a pre-registered margin of 5\pp\ (the declared rule switches from 4 to 5\pp\ when tune discordance exceeds 0.25; measured 0.2629). Price CIs come from only two arrival seeds; we show below that they understate between-run variation by $\approx$1.8$\times$ and treat them as lower bounds on uncertainty.

\textbf{What was pre-registered and what was not.} The claims C1--C4 (identification/detection, load and headroom dependence, joint conditioning, no-gap-no-gain) were fixed before the test read. The Jev results, the RouterBench and HallusionBench analyses, and the SLO simulation were added \emph{after} C1's detection failure and are post hoc.

\section{Results}
\subsection{Root finding: the held-out tier gap is small}
The S1 gate measured a +13.6\pp\ pooled gap between 7B and 32B on a 66-item tune sample and cleared its 8\pp\ bar. On the 626-item held-out test split the paired gap is much smaller (Table~\ref{tab:gap}); only MMLU-Pro excludes zero, and on GPQA the large model is nominally worse. The gate was correct about its own sample but could not distinguish 13\pp\ from 3\pp.

\begin{table}[t]
\centering
\caption{Paired accuracy gap (32B $-$ 7B), held-out test, $n=626$}
\label{tab:gap}
\small
\begin{tabular}{lcc}
\toprule
Dataset & Gap (pp) & 95\% CI \\
\midrule
Pooled & $+2.88$ & $[-0.80,+6.55]$ \\
GPQA-diamond & $-3.76$ & $[-13.53,+6.02]$ \\
HotpotQA & $+2.53$ & $[-3.16,+8.23]$ \\
MATH-500 & $+0.57$ & $[-5.68,+6.82]$ \\
MMLU-Pro & $+11.32$ & $[+3.14,+19.50]$ \\
\bottomrule
\end{tabular}
\end{table}

Everything downstream follows. In the 2,504-row headline stage, \textsc{Triage} and its load-blind variant both score 0.3578 accuracy, identical to always-small, at 4.18 versus 3.77 GPU-s/request: with no benefit to escalating, the allocator does not escalate. A cascade cannot beat a 14B model by escalating to a 32B model that is not better: against a mid-tier baseline (0.3861 at 4.84 GPU-s), the matched-compute cascade scored 0.3974, a paired difference of $+1.13$\pp\ with CI $[-3.20,+5.27]$ ($n=531$), which we report as \textsc{inconclusive}. A random router at the same spend scores 0.4035, so resolving the question would need $n\approx 19{,}600$, more than the mix contains.

\subsection{Evidence corruption: falsified, then repaired post hoc}
On RGB with real tiers, the pre-registered prediction that the reliability state reaches AUROC $\geq 0.70$ is \textbf{falsified}: 0.509 at corruption rate 0.5 and 0.571 at rate 1.0 (Table~\ref{tab:c1}). The prompt-only router scores exactly 0.500, as Proposition~1 requires. Every generation-derived signal is at chance (uncertainty 0.504, instability 0.500, contradiction 0.500): RGB negatives are fluent, on-topic passages, so they do not make the model less certain or less self-consistent. Only signals that compare evidence to the query move at all.

\begin{table}[t]
\centering
\caption{Corruption-detection AUROC on RGB (cluster bootstrap by base item)}
\label{tab:c1}
\small
\setlength{\tabcolsep}{3pt}
\begin{tabular}{lcc}
\toprule
Detector & Rate 0.5 & Rate 1.0 \\
\midrule
Prompt-only (any function of $q$) & 0.500 & 0.500 \\
Lexical reliability state & 0.509 & 0.571 \\
Lexical evidence score & 0.516 & 0.575 \\
Jev ``insufficient'' (served) & 0.573 & \textbf{0.846} \\
\hspace{1em}95\% CI & [0.551, 0.599] & [0.810, 0.881] \\
Jev ``insufficient'' (held-out) & 0.667 & \textbf{0.926} \\
\hspace{1em}95\% CI & [0.620, 0.721] & [0.884, 0.962] \\
Risk with Jev swapped in & 0.537 & 0.695 \\
\bottomrule
\end{tabular}
\end{table}

\begin{figure}[t]
\centering
\begin{tikzpicture}
\begin{axis}[
  ybar, width=\columnwidth, height=4.3cm,
  bar width=9pt, ymin=0.4, ymax=1.0,
  ylabel={AUROC}, ylabel style={font=\scriptsize,yshift=-2pt},
  symbolic x coords={Prompt,Lexical,Jev served,Jev held-out},
  xtick=data, x tick label style={font=\scriptsize},
  y tick label style={font=\scriptsize},
  nodes near coords, nodes near coords style={font=\tiny,/pgf/number format/fixed,/pgf/number format/precision=3},
  every node near coord/.append style={yshift=0pt},
  enlarge x limits=0.15, ymajorgrids, grid style={black!15}
]
\addplot[fill=black!25] coordinates {(Prompt,0.500) (Lexical,0.571) (Jev served,0.846) (Jev held-out,0.926)};
\end{axis}
\end{tikzpicture}
\caption{Detecting fully corrupted RGB evidence. The Jev judge was added after the lexical result was falsified.}
\label{fig:c1}
\end{figure}
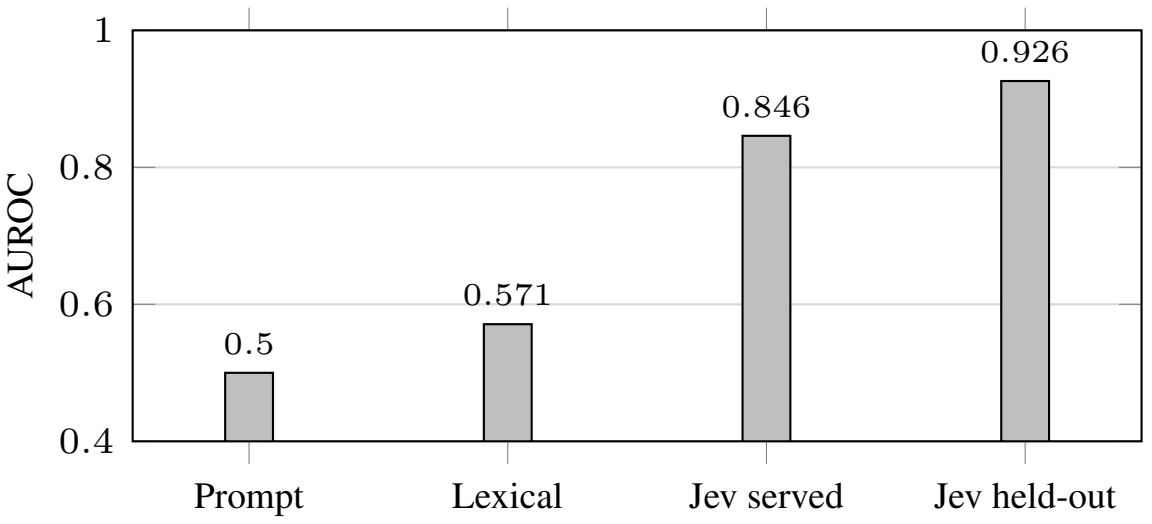

\textbf{Post-hoc hypothesis.} The lexical score is a cosine similarity, and RGB negatives are on-topic, so it cannot separate them. We therefore replay the 600 logged A40 rows with Jev's semantic sufficiency judgment in place of the cosine score; no GPU is used and the abstention weights are \emph{not} retuned. At full corruption Jev reaches 0.846 on served items and 0.926 on the 100 items never served, clearing 0.70 on both. The reconstruction of the logged risk from its components matched on 600/600 rows. At rate 0.5 every signal is weak, Jev included, because half of the passages are still positives, so the label is noisy by construction. Swapping Jev into the existing weighted risk reaches only 0.695, since the evidence term carries 0.15 of the weight and the other four components are at chance; the pre-registered claim about the \emph{whole state} stays falsified, and we claim a better \emph{signal}, not a better state.

\subsection{Predicting wrong answers}
The decision a router needs is whether the answer about to be served is wrong. Treating Jev's ``not supported'' as the score, AUROC on all 600 RGB rows is 0.922 (95\% CI [0.891, 0.951]), against 0.561 [0.460, 0.664] for the logged risk. By subset it is 0.971 on clean, 0.974 at rate 0.5, and 0.762 at rate 1.0 (point estimates, not yet bootstrapped).

\textbf{Caveat that matters.} These RGB answers were generated \emph{closed-book}: the answer prompt contained only the question, and retrieved passages reached the answer only if a verification step ran. The 7B tier was therefore wrong on 71\% of clean items, whose passages contain the gold answer, and the overall error rate is 72.8\%. The grader was checked and is not the cause. Two consequences follow. First, this experiment measures \emph{detection} of corrupted evidence and of unsupported answers, not the \emph{effect} of corruption on answers (quality moves only from 0.290 to 0.255 across rates). Second, a high AUROC at a 73\% base rate is partly the judge noticing that an answer does not match the passages; it is not evidence that routing on this signal would raise accuracy. How many answers a Jev-triggered verification or escalation \emph{repairs} requires re-serving on a GPU and is not yet measured.

\subsection{Three more model pairs on RouterBench}
Costs in this section are API dollars (RouterBench's), not GPU-seconds. Policies escalate a fraction chosen on the tune split. \emph{Prompt-only} has Jev read the question and pays for small \emph{or} big; \emph{triage} has Jev read the small model's answer, always pays for small, and pays for big when escalating. Both are charged for their Jev call ($\approx$\$0.00002).

\begin{table*}[t]
\centering
\caption{RouterBench, test split $n=1{,}708$ per pair. AUROC is for predicting that the small model is wrong; ``1\pp\ point'' is the escalation rate at which triage is within 1\pp\ of always-big (point estimate). $\Delta$ is triage minus the stated baseline at a 50\% budget (paired bootstrap 95\% CI).}
\label{tab:rb}
\small
\setlength{\tabcolsep}{1.6pt}
\footnotesize
\begin{tabular}{lcccccc}
\toprule
Pair (small $\to$ big) & Quality (s/b) & AUROC trg/prompt & $\Delta$ vs prompt (pp) & $\Delta$ vs random (pp) & 1\pp\ point & Cost vs big \\
\midrule
Mistral-7B $\to$ Mixtral-8x7B & 0.360/0.595 & 0.955/0.599 & $+7.67$ [$+5.8,+9.4$] & $-2.47$ [$-4.1,-0.9$] & 44\% esc. & 4.7\% lower \\
WizardLM-13B $\to$ Yi-34B & 0.440/0.649 & 0.945/0.668 & $+2.00$ [$+0.2,+3.9$] & $-2.09$ [$-3.7,-0.5$] & 34\% esc. & 7.1\% lower \\
GPT-3.5 $\to$ GPT-4 & 0.673/0.827 & 0.898/0.681 & $+4.49$ [$+3.4,+5.7$] & $+8.31$ [$+7.2,+9.6$] & 20\% esc. & 79.5\% lower \\
\bottomrule
\end{tabular}
\end{table*}

Table~\ref{tab:rb} shows three points.
\begin{enumerate}
\item The reliability signal transfers: AUROC 0.90--0.96 against 0.60--0.68 for prompt-only. Per task family it is 0.85--0.99, except MBPP with GPT-3.5 (0.69). On MMLU part of this is detecting format failures.
\item The cost benefit depends on the price ratio. With GPT-4 at roughly 17$\times$ the cost of GPT-3.5, the cascade's extra small call is cheap, and at a 25\% budget triage beats prompt-only by $+9.73$ [$+8.17,+11.57$]\pp. With open pairs at 2--3$\times$, the same overhead makes triage \emph{worse than random} at a 50\% budget.
\item ``Within 1\pp'' is a point estimate. The paired CIs of quality minus always-big are $[-2.52,+1.00]$, $[-2.93,+1.00]$, and $[-2.58,+0.59]$\pp, so non-inferiority is established at about 3\pp, not 1\pp.
\end{enumerate}

\textbf{OPE against ground truth.} Because RouterBench is a full-information table, we log one arm per item under a context-dependent behavior policy, estimate triage's value at three escalation rates, and compare with the truth over 200 replications. All three estimators are unbiased to within 0.35\pp; doubly robust (DR) has RMSE 0.86--1.26\pp, about half of IPS (1.67--2.60\pp), and its 95\% CIs cover 0.98--1.00 of the time, i.e., they are conservative.

\subsection{Images: HallusionBench}
In HallusionBench the same question is asked of an original figure and of human-edited copies, so the text is byte-identical across the label (Proposition~1 applies). The served answer is taken to be the original image's ground truth; the label is whether the image in hand still supports it. A prompt-only detector scores 0.500 by construction. Jev, given the image, reaches AUROC \textbf{0.834} [0.802, 0.862] (cluster bootstrap by pair). It is strongest on tables (0.944), charts (0.928), figures (0.885), and OCR (0.884), and weakest on math (0.699) and illusions (0.719). The label detects ``no longer supported'', not ``was edited'': 293 of 356 edits flip the answer. Jev's image input is labeled a prototype by its vendor.

\subsection{The price of an escalation}
We forced an escalation on every eighth request at three offered loads (2 seeds each) and measured the extra GPU-seconds (Table~\ref{tab:price}). The price rises 21.9$\times$ from 1 to 10 rps, with CIs excluding zero at every rate. The 1-rps interval is as wide as its estimate, so the ratio itself is imprecise; pairing the lower limit at 10 rps with the upper limit at 1 rps gives a conservative ratio of about 11$\times$ (our arithmetic).

\begin{table}[t]
\centering
\caption{Escalation price versus offered load (A40, forced escalation). P95 $\Delta$ is the change in small-tier P95 latency.}
\label{tab:price}
\small
\setlength{\tabcolsep}{3pt}
\begin{tabular}{rccc}
\toprule
Load (rps) & Price (GPU-s) & 95\% CI & P95 $\Delta$ (ms) \\
\midrule
1 & 0.01415 & [0.0006, 0.0277] & $+23.4$ \\
4 & 0.23697 & [0.2090, 0.2650] & $+83.5$ \\
10 & 0.30948 & [0.2962, 0.3228] & $+109.3$ \\
\bottomrule
\end{tabular}
\end{table}

\begin{figure}[t]
\centering
\begin{tikzpicture}
\begin{axis}[
  width=\columnwidth, height=4.2cm,
  xlabel={Offered load (rps)}, ylabel={Price (GPU-s)},
  xlabel style={font=\scriptsize}, ylabel style={font=\scriptsize,yshift=-2pt},
  tick label style={font=\scriptsize}, xmin=0, xmax=11, ymin=0, ymax=0.36,
  xtick={1,4,10}, ymajorgrids, grid style={black!15}
]
\addplot[mark=*,thick,error bars/.cd,y dir=both,y explicit]
  coordinates {(1,0.01415) +- (0,0.01355) (4,0.23697) +- (0,0.0280) (10,0.30948) +- (0,0.0133)};
\end{axis}
\end{tikzpicture}
\caption{Marginal price of an escalation versus load. Error bars are the 95\% CI half-widths; they are lower bounds on uncertainty (see text).}
\label{fig:price}
\end{figure}
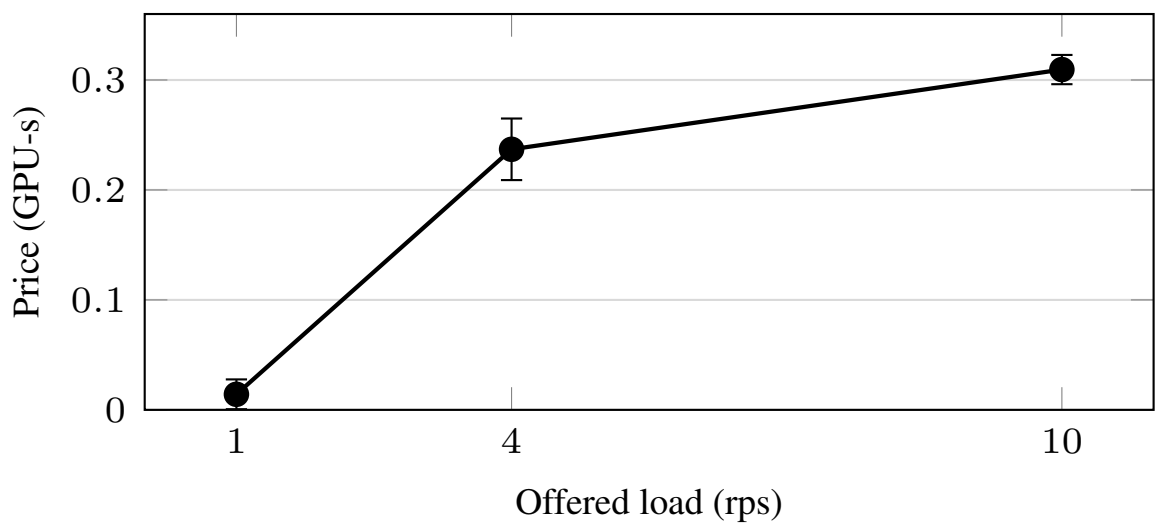

\textbf{Headroom: a replicated null.} Holding load fixed and varying the big tier's KV pool 1,058$\to$2,423 blocks (peak occupancy 0.85$\to$0.37), the price was non-monotone with a total span of about 5\% (0.0354, 0.0337, 0.0355 GPU-s). A full re-run on a fresh host \emph{reversed the sign} of the first step and showed between-run differences about 1.8$\times$ wider than the within-run CI width; the earlier statement that the step was real is withdrawn. The re-run again returned a non-monotone result with a 1.4\% spread across an 8$\times$ pool range. Because the two tiers hold disjoint KV pools, one plausible reading is that the mechanism is compute and bandwidth contention rather than memory, but we did not test that directly. A linear headroom term fit at $R^2=-0.345$, so it was refused.

\textbf{Fitted contention term.} Against measured device concurrency (2.6--82.3; four points), price rose 14.9$\times$ (0.00355$\to$0.0529 GPU-s) and a linear slope of 1.254 fit with $R^2=0.918$. Four points do not validate a functional form; the fit extrapolates to a $13.75\times$ multiplier at the top of its own range.

\subsection{Load-aware versus load-blind allocation}
At the fitted price and an identical $\lambda$ for both arms, the load-aware allocator withdrew from escalation entirely at 2, 6, and 12 rps, while the load-blind arm continued. This cut GPU-seconds by 42.5--85.3\% and P95 latency by 85.2--96.1\%, but \emph{lowered quality by 6.75--7.75\pp}. This is a trade along a frontier, not an iso-compute win, and we do not claim that load-aware allocation ``beats'' load-blind. With an idle device the two policies are identical to the last decimal, because the state multiplier is 1.

Spend is also not monotone in the price of compute: sweeping $\lambda$ from 0.02 to 0.20 \emph{increased} spend by 19\% (1.98$\to$2.36 GPU-s) because information is worth buying only near the decision boundary. The usable range of $\lambda$ on our mix is about 0.02, from 18.3 GPU-s/request (92\% escalated) down to the always-small floor.

\subsection{SLO control in a calibrated simulator}
No GPU was available for a live SLO run, so we calibrated a discrete-event simulator on 4,188 measured A40 calls and declared the SLO (P95 $\leq$ 10\,s) in advance. On five held-out runs it reproduces P95 within $-10\%$ to $+43\%$ and P50 poorly ($-67\%$ to $+75\%$); we treat results as qualitative. At 2--12 rps (Poisson) and on an Azure conversation trace, a controller that only re-prices interventions saturated at its price cap ($13\times$) and still missed the SLO (met 2.5--43\%, versus 0\% for load-blind triage, which escalates 99\% of requests). The binding delay is the small tier's own queue, which no pricing decision touches. Holding a P95 SLO needs admission control or shedding, which exists in the code but is not wired into the arrival runner.

\section{Application Scenarios}
The scenarios below motivated the work. We state, for each, what this paper actually supports.

\begin{table*}[t]
\centering
\caption{Scenarios, supporting evidence, and gaps. None was evaluated end to end on production data.}
\label{tab:use}
\small
\renewcommand{\arraystretch}{1.25}
\begin{tabular}{p{2.4cm}p{5.6cm}p{7.2cm}}
\toprule
Scenario & Evidence in this paper & What is not shown \\
\midrule
Customer-support chatbots & Small-wrong AUROC 0.90--0.96 on five RouterBench task families; cascade within $\approx$3\pp\ of GPT-4 at 79.5\% lower cost (GPT pair). & No support tickets or help-doc grounding. Open-model pairs were below random at a 50\% budget. \\
Search and RAG (enterprise, legal, medical) & Corrupted evidence detected at 0.846/0.926; any prompt-only router is exactly 0.500. & RGB answers were closed-book, so the effect on answer correctness is unmeasured; no legal or medical corpus. \\
Document and invoice processing & Unsupported answers over edited images at 0.834; 0.88--0.94 on tables, charts, OCR. & HallusionBench is not invoices; no field extraction experiment; image input is a prototype. \\
Coding and analytics copilots & GPT-3.5$\to$GPT-4 cost result above. & MBPP is the weakest family (AUROC 0.69 for GPT-3.5); no copilot workload or latency study. \\
Serving under peak load & Escalation price rises 21.9$\times$ (1$\to$10 rps); load-aware allocator withdraws from escalation. & Withdrawal costs 6.75--7.75\pp\ quality; SLO result is simulated and not met without admission control. \\
\bottomrule
\end{tabular}
\end{table*}

\section{Threats to Validity and Limitations}
\textbf{Small effective tier gap.} The 7B$\to$32B difference is not distinguishable from zero pooled, so end-to-end savings on the primary mix are not demonstrated. A model pair with a larger, verified gap is needed.

\textbf{Post-hoc signal.} Jev results come after a falsified prediction. The held-out RGB split and the unchanged abstention weights reduce, but do not remove, the risk of fitting to the result; one corpus and one judge version were used.

\textbf{Closed-book answers.} RGB answers did not read the evidence. Detection results stand as detection; repair is untested.

\textbf{Measurement noise.} Price CIs use two seeds and understate between-run variability by about 1.8$\times$. The contention fit has four points. The simulator is coarse.

\textbf{External data and cost units.} RouterBench costs are API dollars; GPU-second savings are not implied. The vendor judge adds a network call (p50 about 0.4 s), about 40\% of a small-tier RGB request if called serially, and its latency must be reported next to any P95 it touches.

\textbf{Incomplete campaign.} The load and robustness stages were not re-run after the pod hit its deadline; 5,033 rows were recovered, including the complete 2,504-row headline. A second device class was not run.

\section{Conclusion}
Reading the evidence, not just the prompt, is what lets a router see corrupted retrieval: prompt-only routers are provably at chance, and a semantic judge detects corruption at AUROC 0.85--0.93 where a lexical score does not. Across three RouterBench pairs the post-generation signal predicts small-model errors at 0.90--0.96, and it pays off in cost only when the large model is much more expensive than the small one. Escalation on a shared GPU becomes markedly more expensive with load, but not with KV headroom. We could not show an end-to-end serving win, because the large model was not reliably better, and a short gate hid that. Next steps are re-serving RGB with evidence in the answer prompt to measure repair, a live SLO run with admission control, and a model pair with a verified quality gap.

\textbf{Reproducibility.} Code, run directories, and cached judgments will be released in public repository. Offline scripts regenerate the Jev and RouterBench tables from cached judgments. Raw served logs are withheld because the GPQA terms forbid redistribution.

\bibliography{refs}
\bibliographystyle{mlsys2025}

\end{document}